\documentclass[10pt,twocolumn,letterpaper]{article}

\usepackage{cvpr}              

\usepackage{graphicx}
\usepackage{amsmath}
\usepackage{amssymb}
\usepackage{booktabs}
\usepackage{multirow}
\usepackage{algorithm}
\usepackage{algorithmic}

\usepackage[pagebackref,breaklinks,colorlinks]{hyperref}

\usepackage[capitalize]{cleveref}
\crefname{section}{Sec.}{Secs.}
\Crefname{section}{Section}{Sections}
\Crefname{table}{Table}{Tables}
\crefname{table}{Tab.}{Tabs.}

\def\cvprPaperID{2} 
\def\confName{CVPR}
\def\confYear{2023}

\begin{document}
	
	\title{Explore the Power of Synthetic Data on Few-shot Object Detection} 

\author{Shaobo Lin, Kun Wang, Xingyu Zeng, Rui Zhao\\
	Sensetime Research\\
	{\tt\small $\{$linshaobo,wangkun,zengxingyu,zhaorui$\}$@sensetime.com}
}
\maketitle

\begin{abstract}
Few-shot object detection (FSOD) aims to expand an object detector for novel categories given only a few instances for training. The few training samples restrict the performance of FSOD model. Recent text-to-image generation models have shown promising results in generating high-quality images. How applicable these synthetic images are for FSOD tasks remains under-explored. This work extensively studies how synthetic images generated from state-of-the-art text-to-image generators benefit FSOD tasks. We focus on two perspectives: (1) How to use synthetic data for FSOD? (2) How to find representative samples from the large-scale synthetic dataset? We design a copy-paste-based pipeline for using synthetic data. Specifically, saliency object detection is applied to the original generated image, and the minimum enclosing box is used for cropping the main object based on the saliency map. After that, the cropped object is randomly pasted on the image, which comes from the base dataset. We also study the influence of the input text of text-to-image generator and the number of synthetic images used. To construct a representative synthetic training dataset, we maximize the diversity of the selected images via a sample-based and cluster-based method. However, the severe problem of high false positives (FP) ratio of novel categories in FSOD can not be solved by using synthetic data. We propose integrating CLIP, a zero-shot recognition model, into the FSOD pipeline, which can filter 90$\%$ of FP by defining a threshold for the similarity score between the detected object and the text of the predicted category. Extensive experiments on PASCAL VOC and MS COCO validate the effectiveness of our method, in which performance gain is up to 21.9$\%$ compared to the few-shot baseline.

\end{abstract}

\section{Introduction}
\label{sec:intro}

In recent years, we have tremendous progress in object detection~\cite{ren2016faster,redmon2017yolo9000,lin2017feature,carion2020end,dai2021dynamic}. However, the impressive performance of these models relies on a large amount of annotated data. Therefore, the detectors cannot generalize well to novel categories, especially when the annotated data are scarce. In contrast, humans can learn to recognize or detect a novel object with only a few labeled examples. Few-shot object detection (FSOD), which simulates this way, has attracted increasing attention. In FSOD, an object detector that is trained using base categories with sufﬁcient data (base images) can learn to detect novel categories using only a few annotated images (novel images).

\begin{figure}[t]
\vskip -0.1in
\begin{center}
	\centerline{\includegraphics[width=\columnwidth]{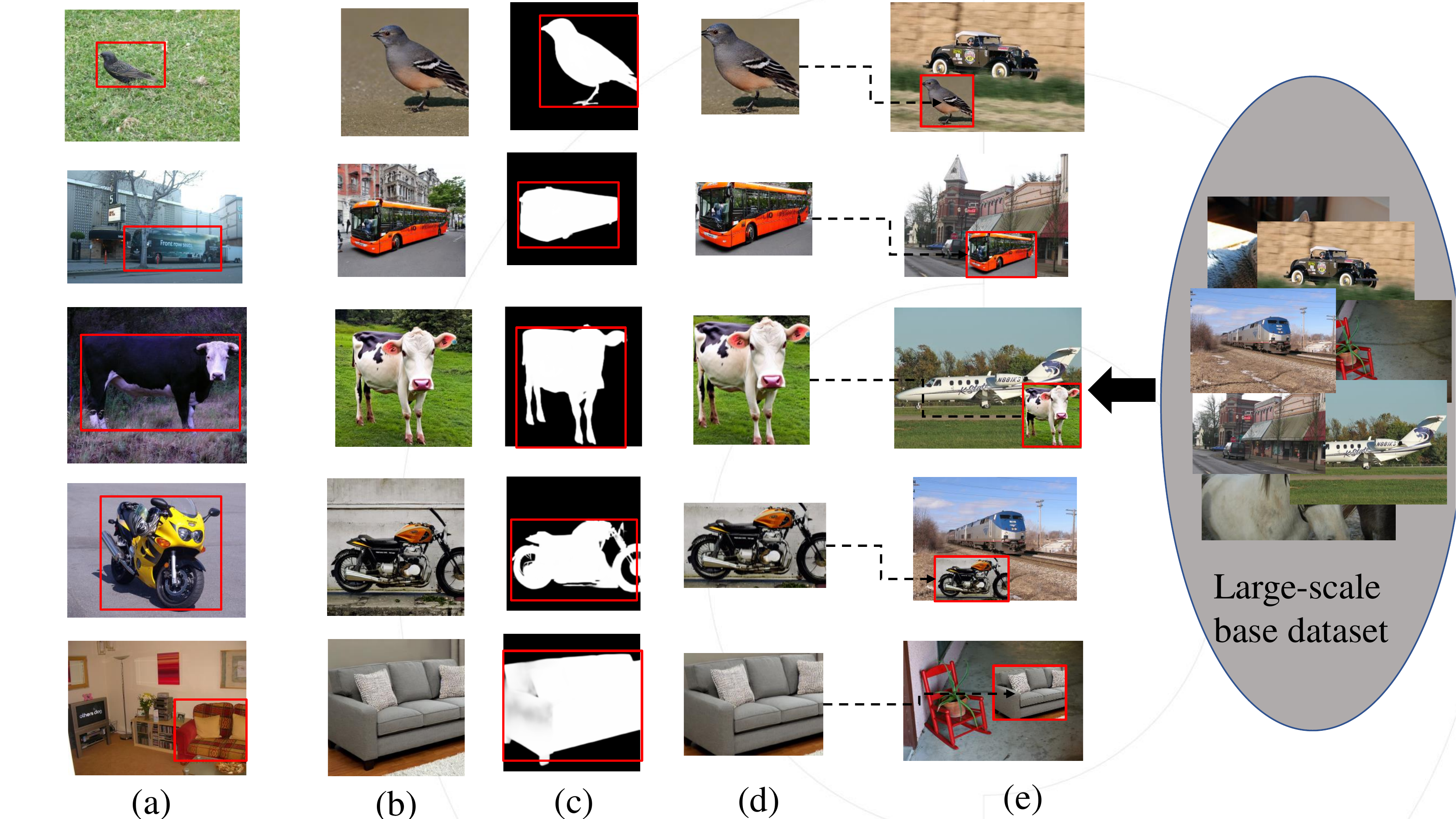}}
	\caption{How to utilize synthetic data for FSOD? (a) The few-shot novel categories, including bird, bus, cow, motorbike, and sofa, are from the dataset of PASCAL VOC split 1. (b) The synthetic data from Stable Diffusion. (c) The saliency detection map of the synthetic data. (d) The cropped novel instances are based on the minimum enclosing box of the saliency map. (e) Combining synthetic instances and the randomly selected background from the base dataset.}
	\label{fig0}
\end{center}
\vskip -0.3in
\end{figure}

The few training samples restrict the performance of FSOD model.  In FSOD, the number of objects for each novel category is $K$ for $K$-shot detection. Existing literature always controls the number of novel data ($K$-shot), to carefully compare the performance of different methods. While this is helpful to drive research on sophisticated algorithms, it is unhelpful for the industry to enhance its few-shot based applications by incorporating external synthetic data. 
With the recent development of generative models, the text-to-image generator has made a great process. For example, DALL.E~\cite{ramesh2022hierarchical}, Imagen~\cite{saharia2022photorealistic}, and Stable Diffusion~\cite{rombach2022high} can generate high-quality images by simply using the input text description. These generators can produce diverse outcomes, implying a brighter future for industrial applications such as tackling many existing few-shot or long-tailed issues. 
This encourages us to explore the impact of synthetic data from text-to-image generators on FSOD tasks. Our study is carried out on the open-sourced Stable Diffusion. By using the generated images, we define a new setting of $(K+G)$-shot for the problem of using synthetic data for few-shot learning, consisting of $K$ real novel instances and $G$ generated novel instances.  Two key questions must be answered: (1) How to use synthetic data for FSOD? (2) How to find representative samples from the large-scale synthetic dataset? 


{\bf How to use synthetic data for FSOD?} In Figure~\ref{fig0}, the original FSOD dataset from PASCAL VOC split1 is annotated with the bounding boxes and categories. Since the synthetic images from Stable Diffusion lack annotations, we cannot utilize these data directly. Copy-Paste~\cite{park2022majority,ghiasi2021simple,lin2023effective} is a simple and effective data augmentation strategy. Randomly pasting generated novel instances onto background images can generate a combinatorial number of training data for free. However, the original generated images can't be pasted onto the background image. The reason is that using the generated images, including the non-negligible background regions and the inaccurate bounding boxes, leads to performance degradation shown in Table~\ref{table:3}. To solve this problem, saliency object detection is applied to the original generated images, and the minimum enclosing box is used for cropping the main object. After that, the cropped object is randomly down-sized and pasted on the image randomly selected from the base dataset. Besides, we design and compare several input text formats, finding that the performance of different text formats is related to the few-shot settings, shown in Figure~\ref{fig:prompt}. Intuitively, FSOD can benefit from more training data by increasing $G$ in the $(K+G)$-shot setting. However, using more synthetic data can not continuously improve the FSOD model in our work. We find that using 20 synthetic images works best in most few-shot cases. 

Since we can obtain a large amount of generated images from a text-to-image generator, {\bf finding the representative samples} is necessary. Authenticity and diversity are two factors for evaluating the quality of a generated dataset. We focus on maximizing the diversity of training dataset for selecting the representative synthetic data because the authenticity of the images from Stable Diffusion is remarkable. We introduce the sample-based and cluster-based (k-means~\cite{hartigan1979algorithm} / spectral clustering~\cite{ng2001spectral}) methods to build a high-diversity dataset. Specifically, the sample-based method applies uniform sampling to the data feature of each category. The cluster-based method uses a clustering algorithm to separate the data feature of each category into several sub-spaces and select one sample in each sub-space to maximize the diversity of training dataset.
We select the sample nearest to the cluster centroid for each sub-space. CLIP~\cite{radford2021learning} is used as the feature extractor for getting the data feature. 
However, the FP of novel categories is prominent, in which the base categories are often recognized as novel ones. The problem of high FP ratio of novel categories can not be solved by synthetic data, shown in Figure~\ref{fig:CLIP-FP}. CLIP is a zero-shot model that uses natural language to retrieve related images. We propose to integrate CLIP into the FSOD model for filtering the FP. By computing the similarity score between the detected object from the FSOD model and the name list of categories, we can remove the proposals whose similarity score with the predicted category is lower than the pre-defined threshold. 

Our key contributions can be summarized as: (1) To the best of our knowledge, we are the first to leverage the external synthetic novel data from a text-to-image generator for FSOD, showing significant application potential. (2) We explore how to utilize the synthetic novel data, including the copy-paste pipeline, the design of input text, the number of used synthetic images, and the data selection strategy for discovering the representative synthetic samples.  
(3) Our method significantly improves multiple baselines and achieves state-of-the-art performance on PASCAL VOC and MS COCO. 


\begin{figure*}[t]
\begin{center}
	\centerline{\includegraphics[width=16cm]{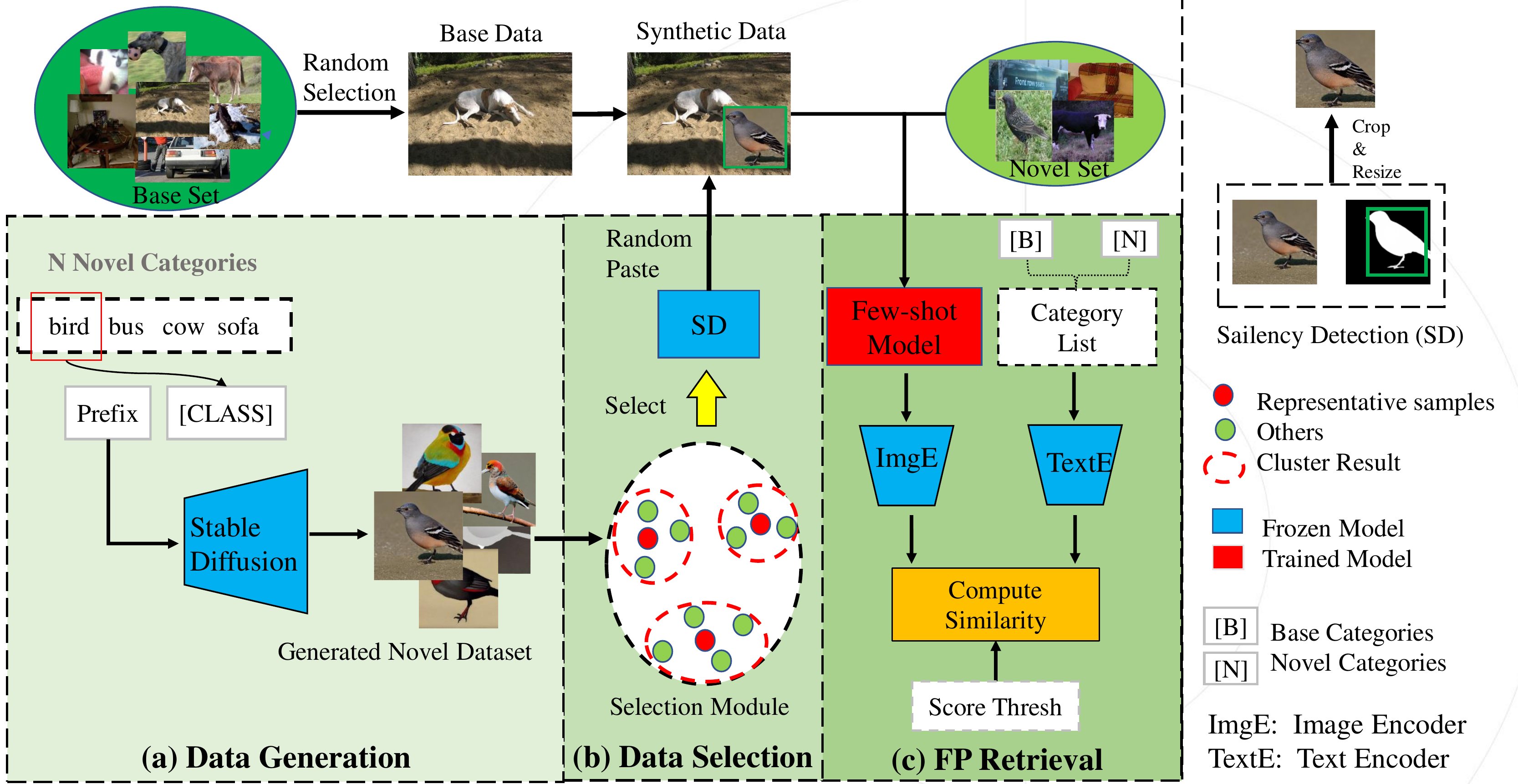}}
	\caption{The pipeline of using synthetic data for FSOD. (a) During the data generation, the novel dataset is generated by Stable Diffusion. (b) In the data selection step, the generated data is processed by the selection module to find representative samples for training. We use the cluster-based method as an example. After that, a saliency detection is applied to the selected samples for cropping the novel instances. The synthetic images, added to the few-shot set for training the few-shot model, are obtained by randomly pasting novel instances onto the randomly selected background images. The Few-shot model can be any current few-shot object detector. (c) For FP retrieval, the similarity of the object feature from the image encoder of CLIP and the text feature from the text encoder of CLIP is computed. The score thresh filters the FP based on the similarity score.}
	\label{fig2}
\end{center}
\vskip -0.4in
\end{figure*}

\section{Related Works}
\subsection{Few-shot Object Detection}

FSOD is an important yet unsolved task in computer vision. Some works use meta-learning~\cite{fan2020few,kang2019few,yan2019meta,wang2019meta,li2021beyond}, where a meta-learner is introduced to acquire class-agnostic meta-knowledge which is transferred to novel classes. 
With the help of a meta learner that takes the support images as well as the bounding box annotations as inputs, the feature re-weighting modules are applied to a single-stage object detector (YOLOv2)~\cite{kang2019few} and a two-stage object detector (Faster R-CNN)~\cite{yan2019meta}. 
\cite{li2021beyond} uses a class margin equilibrium (CME) approach, with the aim to optimize both feature space partition and novel class reconstruction in a systematic way. Transformation invariant principle ~\cite{li2021transformation} is proposed for various meta-learning models by introducing consistency regularization on predictions from the transformed images.
TFA~\cite{wang2020frustratingly} is a simple two-stage fine-tuning approach, which significantly outperforms the earlier meta-learning methods. 
MPSR~\cite{wu2020multi} adopts multi-scale positive sample refinement to handle scale variance problem. 
FSCE~\cite{sun2021fsce} aims to learn contrastive-aware object proposal encodings that facilitate the classification of detected objects. FADI~\cite{cao2021few} uses a two-step fine-tuning framework via association and discrimination, which builds up a discriminative feature space for each novel class with two integral steps. 
DeFRCN~\cite{qiao2021defrcn} extends Faster
R-CNN by using gradient decoupled layer for multi-stage decoupling and prototypical calibration block for multi-task decoupling. 
Pseudo-Labelling~\cite{kaul2022label} is proposed to find previously unlabelled instances to obtain high-quality pseudo-annotations for novel categories from the training dataset. 
MFDC~\cite{wu2022multi} proposes to learn three
types of class-agnostic commonalities between base and novel classes: recognition-related, localization-related and distribution commonalities, which are able
to be integrated into most of existing fine-tuning based methods by performing distillation via a unified distillation framework. 


\subsection{Text-to-Image Generators}

Recently, diffusion models~\cite{song2020improved,ho2020denoising} have become a promising generative modeling framework, achieving state-of-the-art performance on image generation tasks~\cite{dhariwal2021diffusion,ho2022cascaded}. DALL.E~\cite{ramesh2022hierarchical} proposes a two-stage model: a prior that generates a CLIP~\cite{radford2021learning}
image embedding given a text caption and a decoder that generates an image based on the image embedding.
Imagen~\cite{saharia2022photorealistic} is a text-to-image diffusion model based on the power of language models and diffusion models. 
Using cross-attention layers, Stable Diffusion~\cite{rombach2022high} uses diffusion models to synthesize high-resolution images for general conditioning inputs. 
GLIDE~\cite{nichol2021glide} studies diffusion models for the text-conditional image synthesis by comparing two guidance strategies: CLIP and classifier-free guidance. The synthetic images generated from GLIDE
are used for zero-shot, few-shot, and regular image recognition tasks~\cite{he2022synthetic}, showing the power of synthetic data on image recognition. However,~\cite{he2022synthetic} does not consider the influence of the number of synthetic data defined by $G$ in the $(K+G)$-shot setting.
By comparing Stable Diffusion to GLIDE in Table~\ref{table:gen}, we notice that the quality of synthetic data from Stable Diffusion is superior to that of GLIDE. Therefore, we use Stable Diffusion in our experiments. 

\section{Method}

\subsection{Preliminary}
In FSOD, given a labeled base dataset $D_B$, there are $C_B$ base classes with sufficient images in each class. Novel dataset $D_N$ with novel classes $C_N$ consists of a few samples in each class. $C_B$ and $C_N$ do not have overlapping categories. The number of instances for each class in $C_N$ is $K$ for $K$-shot detection. Using the generated images, we define a new setting of $(K+G)$-shot for the problem of using synthetic data for FSOD, which consists of $K$ real novel instances and $G$ generated novel instances. There are two stages in FSOD methods~\cite{wang2020frustratingly,sun2021fsce,li2021beyond}. In the pre-training stage, the model is trained on base classes to obtain a robust feature representation. In the fine-tuning stage, the pre-trained model is then fine-tuned on a balanced few-shot set (or adding the synthetic novel data in our setting), including both base and novel classes ($C_B \cup C_N$).

\subsection{Our Pipeline}

In Figure~\ref{fig2}, there are three steps in our overall pipeline of using synthetic data for FSOD: data generation, data selection, and FP retrieval.

{\bf Data Generation} 
To study the influence of the input text of text-to-image generator, several representative text descriptions for Stable Diffusion are designed by following the format of prefix with the name of a category, shown in Table~\ref{table:31}. Specifically, We can simply use the class name as the input text whose prefix is "None", or "a / one" with the class name. Besides, we can combine "a / one" and "photo / picture" or use some additional representative adjectives as the prefix of input text. The comparison among these text formats will be discussed in section~\ref{sec:G}.
Using the input text, we can get many synthetic images according to the category name. Because the entity of the input text of Stable Diffusion is the name of a category, most of the generated images only have one target.

\begin{table}[t]
\scriptsize
\setlength{\tabcolsep}{6pt}		
\begin{center}
	\caption{The design of prefixes in the input text, including seven different settings: None, a, one, a5, one5, real, and adj.}
	\label{table:31}
	\begin{tabular}{cccc}
		\hline\noalign{\smallskip}
		Methods& Prefixs &Methods& Prefixs\\
		\noalign{\smallskip}
		\hline
		\noalign{\smallskip}
		None&&&\\
		\hline
		\noalign{\smallskip}
		a& "a"&one& "one"\\  
		\hline
		\noalign{\smallskip}        
		\multirow{5}*{a5}& "a"&\multirow{5}*{one5}&'one'\\
		&'a photo of '&&'a photo of one '\\
		& 'a photo of a '&&'a picture of one '\\ 
		&'a picture of '&&'one photo of '\\
		&'a picture of a '&&'one picture of '\\  
		\hline
		\noalign{\smallskip}         
		\multirow{5}*{real}&'real '&\multirow{5}*{adj}&'a photo of a good '\\ 
		&'a real '&&'a photo of a large '\\
		&'one real '&&'a photo of a nice '\\
		&'a photo of a real '&&'a photo of a cool '\\
		&'a photo of one real '&&'a photo of a clean '\\     
		\hline
	\end{tabular}
\end{center}
\vskip -0.2in
\end{table}

{\bf Data Selection} The generated novel data from Stable Diffusion is processed by the selection module to find the representative synthetic samples, which should be high-quality and high-diversity. We focus on maximizing the diversity of the selected data because the quality of the synthetic images is high enough. The selection module is based on the CLIP similarity score or image feature from CLIP. We propose two types of selection methods, including sample-based and cluster-based methods.

\begin{equation}
G_n=\{G_{n}^{i}\}_{i=1}^{N},\  R_n=\{R_{n}^{i}\}_{i=1}^{M}
\end{equation}
where $G_n$ includes the generated novel instances of category $n$. $G_{n}^{i}$ is the $i$-th generated sample of category $n$. $R_n$ is the real novel data of category $n$ from the few-shot dataset. $R_{n}^{i}$ is the $i$-th real sample of category $n$. $N$ and $M$ are the numbers of generated and real novel images, respectively.

The sample-based method is directly based on the CLIP score or image feature of CLIP. As illustrated in Equ \ref{equ:2}, the diversity of the generated dataset can be measured by the CLIP score of the generated instances or the distance between the CLIP features of generated and real novel instances. High-diversity means using uniform sampling to maximize the difference among the selected samples.

\begin{equation}
\label{equ:2}
\begin{split}
	R_s = \left\{
	\begin{array}{l}
		Se(CLIP(G_n,T)) \\
		Se(cos(CLIP(G_n),CLIP(R_n)))
	\end{array}
	\right.
\end{split}
\end{equation}
where $T$ is the input text. By inputting the text and image, $CLIP$ outputs the similarity score. If $T$ is not used, $CLIP$ means the image encoder of CLIP, which processes the generated and real novel images for obtaining the image features. $Se$ is the selection module for finding the representative samples using uniform sampling. The distance between the generated images and real images is computed by the cosine similarity. $R_s$ is the selected samples based on the sample-based methods. 

Using a clustering algorithm is the second way to find the representative samples. 


\begin{equation}
C_n = Clu(CLIP(I_n),k_n)
\end{equation}
where $C_n$ is the class center of category $n$. $Clu$ is the clustering method. $k_n$ is the number of centers of category $n$. We use k-means or spectral clustering to compute the class center. We find that using the spectral clustering algorithm achieves the best performance.

\begin{equation}
R_c = \arg \underset{i}{min}\ cos(C_n,G_{n}^{i})_{i=1}^{N}
\end{equation}

We select the sample nearest to the $k_n$ cluster centroids $C_n$ as the representative samples for category $n$. The distance is computed by the cosine similarity. $R_c$ is the samples selected by the cluster-based methods.

After finding the representative samples, we apply a saliency detection method to these generated images. Then, the minimum enclosing box is used for cropping the main object based on the saliency map. Finally, the cropped object is down-sized and pasted on the randomly selected background from the base dataset. 
These synthetic novel images are introduced into the original few-shot set to obtain a new few-shot training dataset. The number of the selected synthetic images, $G$ in the setting of $(K+G)$-shot, is a crucial hyper-parameter for FSOD, which will be discussed in section ~\ref{sec:G}. 

{\bf FP Retrieval}
In our work, the FP of novel categories is prominent. However, the problem of high FP ratio can not be solved by introducing synthetic data, which is shown in Figure~\ref{fig:CLIP-FP} in section \ref{sec:CLIP}. With the increase of generated images, the FP ratio of the original few-shot model ($w/o$ CLIP) does not decrease. 
We propose integrating the CLIP into the FSOD model for filtering the FP. Formally, given $I_n$ which is the few-shot training dataset of the novel category $n$, we use the FSOD model $Fs$ to get the bounding box $bb_n$ and the corresponding confidence score $conf_n$. 

\begin{equation}
R_n = I_n[bb_n]; conf_n, bb_n = Fs(I_n)
\end{equation}
where $R_n$ is the cropped region of $I_n$ based on $bb_n$, which is predicted to category $n$ by $Fs$.

\begin{equation}
score = \frac{e^{(cos(ImgE(R_n),TextE(t_n)))}}{ \sum_{i = 1}^{L} e^{(cos(ImgE(R_n),TextE(t_i)))}} 
\end{equation}
where $ImgE$ and $TextE$ are the image encoder and text encoder of CLIP. $ImgE$ gets the image feature of the detected object $R_n$. $TextE$ gets the text feature of the input text $t_n$ of the novel category $n$. We compute the cosine similarity of the image and text features. The input text $t_i$ $\in$ $(B\cup N)$ or $N$ means the base and novel categories or only novel categories. $L$ is the length of the category list. In Table~\ref{table:CLIP-use}, we find that the selection of the category list is related to the scale of the dataset, i.e., PASCAL VOC and MS COCO. Finally, we use the softmax operation to get the normalized similarity $score$. We remove the detection results of $R_n$ (FP) whose $score$ is lower than the pre-defined CLIP threshold, which is 0.1 in our experiment. The filtered results are used for final evaluation. With CLIP, we can reduce the FP ratio up to 90 $\%$ shown in Figure~\ref{fig:CLIP-FP}. 

\section{Experiments}

\subsection{Datasets and Evaluation Protocols}

We evaluate our methods on PASCAL VOC~\cite{everingham2010pascal,everingham2015pascal} and MS COCO~\cite{lin2014microsoft}. In PASCAL VOC, we adopt the common strategy~\cite{ren2016faster,redmon2017yolo9000}  that using VOC 2007 test set for evaluating while VOC 2007 and 2012 train/val sets are used for training. Following~\cite{yan2019meta}, 5 out of its 20 object categories are selected as the novel classes, while the remaining
as the base classes. We evaluate with three different novel/base splits from~\cite{yan2019meta}, named as split 1, split 2 and split 3.
Each split contains 15 base categories with abundant data and 5 novel categories with K annotated instances for K = 1, 3, 5, 10. Following~\cite{yan2019meta,wang2020frustratingly,sun2021fsce},  we use the mean average precision (mAP) of novel categories at 0.5 IoU threshold as the evaluation metric and report the results on the official test set of VOC 2007. When using MS COCO, 20 out of 80 categories are reserved as novel classes, the rest 60 categories are used as base classes. The detection performance with COCO-style AP and AP$_{75}$ for K = 1, 3, 5, 10, 30 shots of novel categories are reported.

\setlength{\tabcolsep}{8.5pt}
\begin{table*}[t]
\scriptsize

\begin{center}
	\caption{Comparison with state-of-the-art few-shot object detection methods on VOC2007 test set for novel classes of the three splits. {\bf Black} indicates state-of-the-art. {\color{red}Red} is the improvement compared to the baseline. * means using our methods: the synthetic images, the selection strategy, and the CLIP model for filtering the FP.}
	\label{table:1}
	\begin{tabular}{llcccccccccccc}
		\hline\noalign{\smallskip}
		&&\multicolumn{4}{c}{split 1} & 	\multicolumn{4}{c}{split 2}&\multicolumn{4}{c}{split 3}\\
		\hline\noalign{\smallskip}
		\multicolumn{2}{c}{Methods / Shots}& 1 & 3 &5& 10 & 1 &3&5&10&1&3&5&10\\
		\noalign{\smallskip}
		\hline
		\noalign{\smallskip}
		FRCN+ft~\cite{yan2019meta}&ICCV2019&11.9&29&36.9&36.9&5.9&23.4&29.1&28.8&5.0&18.1&30.8&43.4\\
		FRCN+ft-full~\cite{yan2019meta}&ICCV2019&13.8&32.8&41.5&45.6&7.9&26.2&31.6&39.1&9.8&19.1&35&45.1\\
		FR~\cite{kang2019few}&ICCV2019&14.8&26.7&33.9&47.2&15.7&22.7&30.1&40.5&21.3&28.4&42.8&45.9\\
		MetaDet~\cite{wang2019meta}&ICCV2019&18.9&30.2&36.8&49.6&21.8&27.8&31.7&43&20.6&29.4&43.9&44.1\\
		Meta R-CNN~\cite{yan2019meta}&ICCV2019&19.9&35&45.7&51.5&10.4&29.6&34.8&45.4&14.3&27.5&41.2&48.1\\
		TFA~\cite{wang2020frustratingly}&ICML2020&39.8&44.7&55.7&56.0&23.5 &34.1 &35.1 &39.1& 30.8 & 42.8& 49.5 &49.8\\
		MPSR~\cite{wu2020multi}&ECCV2020&41.7&51.4&55.2&61.8&24.4&39.2&39.9&47.8&35.6&42.3&48&49.7\\
		CME~\cite{li2021beyond}&CVPR2021&41.5&50.4&58.2&60.9&27.2&41.4&42.5&46.8&34.3&45.1&48.3&51.5\\
		FSCN~\cite{li2021few}&CVPR2021&40.7&46.5&57.4&62.4&27.3&40.8&42.7&46.3&31.2&43.7&50.1&55.6\\	
		HallucFsDet~\cite{zhang2021hallucination}&CVPR2021&47&46.5&54.7&54.7&26.3&37.4&37.4&41.2&40.4&43.3&51.4&49.6\\
		FSCE~\cite{sun2021fsce}&CVPR2021 &44.2&51.4&61.9&63.4&27.3&43.5&44.2&50.2&37.2&47.5&54.6&58.5\\
		UPE~\cite{wu2021universal}&ICCV2021&43.8&50.3 &55.4& 61.7& 31.2 & 41.2&42.2 & 48.3& 35.5 & 43.9 &50.6& 53.5 \\
		QA-FewDet~\cite{han2021query}&ICCV2021&42.4&55.7&62.6&63.4&25.9&46.6&48.9&51.1&35.2&47.8&54.8&53.5\\
		Meta faster-rcnn~\cite{han2021meta}&AAAI2021&43&60.6&66.1&65.4&27.7&46.1&47.8&51.4&40.6&53.4&59.9&58.6\\
		FADI~\cite{cao2021few}&NIPS2021&50.3&54.2&59.3&63.2&30.6&40.3&42.8&48&45.7&49.1&55&59.6\\	
		DeFRCN~\cite{qiao2021defrcn}&ICCV2021&53.6&61.5&64.1&60.8&30.1&47.0&53.3&47.9&48.4&52.3&54.9&57.4\\
		FCT~\cite{han2022few}&CVPR2022&49.9&57.9&63.2&67.1&27.6&43.7&49.2&51.2&39.5&52.3&57&58.7\\
		KFSOD~\cite{zhang2022kernelized}&CVPR2022&44.6&54.4&60.9&65.8&37.8&43.1&48.1&50.4&34.8&44.1&52.7&53.9\\	
		Pseudo-Labelling~\cite{kaul2022label}&CVPR2022&54.5& 58.8& 63.2& 65.7& 32.8& 50.7& 49.8& 50.6& 48.4 & 55.0& 59.6 &59.6\\
		MFDC~\cite{wu2022multi}&ECCV2022&63.4&67.7&69.4&68.1&42.1&53.4&55.3&53.8&56.1&59.0&62.2&63.7\\
		\hline
		\noalign{\smallskip}
		\multirow{2}*{DeFRCN*}&Ours& 67.5&{\bf 69.8}&{\bf71.1}&71.5&{\bf52}&54.3&57.5&{\bf57.4}&55.9&58.6&59.6&63.9\\
		&Improve&{\color{red}+13.9}&{\color{red}+8.3}&{\color{red}+7}&{\color{red}+10.7}&{\color{red}+21.9}&{\color{red}+7.3}&{\color{red}+4.2}&{\color{red}+9.5}&{\color{red}+7.5}&{\color{red}+6.3}&{\color{red}+4.7}&{\color{red}+6.5}\\
		\multirow{2}*{MFDC*}&Ours&{\bf68.9}&69.5&70.9&{\bf73.3}&50.5&{\bf56.8}&{\bf58.9}&55.6&{\bf59.9}&{\bf62.8}&{\bf65.3}&{\bf66.9}\\
		&Improve&{\color{red}+5.4}&{\color{red}+1.8}&{\color{red}+1.5}&{\color{red}+5.2}&{\color{red}+8.4}&{\color{red}+3.4}&{\color{red}+3.6}&{\color{red}+1.8}&{\color{red}+3.8}&{\color{red}+3.8}&{\color{red}+3.1}&{\color{red}+3.2}\\			
		
		\hline	
	\end{tabular}
\end{center}
\vskip -0.2in
\end{table*}

\setlength{\tabcolsep}{10pt}	
\begin{table*}[t]
\scriptsize
\begin{center}
	\caption{Few-shot object detection performance on MS COCO. {\bf Black} indicates the state-of-the-art. {\color{red}Red} is the improvement compared to the baseline. * means using our methods: the synthetic images, the selection strategy, and the CLIP model for filtering the FP.}
	\label{table:2}
	\begin{tabular}{lcccccccccc}
		\hline\noalign{\smallskip}
		Methods / Shots& \multicolumn{2}{c}{1} & \multicolumn{2}{c}{3} & \multicolumn{2}{c}{5} & \multicolumn{2}{c}{10} & \multicolumn{2}{c}{30} \\
		\hline\noalign{\smallskip}
		&nAP & nAP$_{75}$&nAP & nAP$_{75}$&nAP & nAP$_{75}$&nAP & nAP$_{75}$&nAP & nAP$_{75}$\\
		\noalign{\smallskip}
		\hline
		\noalign{\smallskip}
		FR~\cite{kang2019few}&-&-&-&-&-&-&5.6&4.6&9.1&7.6\\
		Meta R-CNN~\cite{yan2019meta} &-&-&-&-&-&-&8.7&6.6&12.4&10.8\\	
		CME~\cite{li2021beyond}&-&-&-&-&-&-&15.1&16.4&16.9&17.8\\
		FSCE~\cite{sun2021fsce}&-&-&-&-&-&-&11.1&9.8&15.3&14.2\\
		UPE~\cite{wu2021universal}&-&-&-&-&-&-&11&10.7&15.6&15.7\\
		TFA~\cite{wang2020frustratingly}&3.4&3.8&6.6&6.5&8.3&8.0&10&9.3&13.7&13.4\\
		MSPR~\cite{wu2020multi}&2.3&2.3&5.2&5.1&6.7&6.4&9.8&9.7&14.1&14.2\\
		QA-FewDet~\cite{han2021query}&4.9&4.4&8.4&7.3&9.7&8.6&10.2&9.0&11.5&10.3\\
		Meta faster-rcnn~\cite{han2021meta}&5.0&4.6&&&12.7&16.6&25.7&31.8&10.8&15.8\\
		FADI~\cite{cao2021few}&5.7&6.0&8.6&8.3&10.1&9.7&12.2&11.9&16.1&15.8\\
		DeFRCN~\cite{qiao2021defrcn}&6.5&6.9&13.4&13.6&15.3&14.6&18.5&17.6&22.6&22.3\\
		FCT~\cite{han2022few}&5.6&-&11.1&-&14.0&-&17.1&17&21.4&22.1\\
		KFSOD~\cite{zhang2022kernelized}&-&-&-&-&-&-&18.5&18.7&-&-\\
		Pseudo-Labelling~\cite{kaul2022label}&-&-&-&-&-&-&17.8& 17.8 &{\bf24.5} & {\bf25.0}\\
		MFDC~\cite{wu2022multi}&10.8&11.6&15&15.5&16.4&17.3&19.4&20.2&22.7&23.2\\
		\hline
		\noalign{\smallskip}			
		DeFRCN*&{\bf 18.1}&{\bf19.5}&{\bf19.2}&{\bf20.2}&{\bf19.8}&{\bf20.4}&{\bf20.7} &{\bf21.3}&23.1&23.9\\
		Improve&{\color{red}+11.6}&{\color{red}+12.6}&{\color{red}+5.8}&{\color{red}+6.6}&{\color{red}+4.5}&{\color{red}+5.8}&{\color{red}+2.2}&{\color{red}+3.7}&{\color{red}+0.5}&{\color{red}+1.6}\\
		MFDC*&{\bf 18.1}&19&18.6&19.4&19.5&20.1&20.3&20.9&22.5&23\\
		Improve&{\color{red}+7.3}&{\color{red}+7.4}&{\color{red}+3.6}&{\color{red}+3.9}&{\color{red}+3.1}&{\color{red}+2.8}&{\color{red}+0.9}&{\color{red}+0.7}&-0.2&-0.2\\
		\hline
		
	\end{tabular}
\end{center}
\vskip -0.2in
\end{table*}

\begin{table}[t]
\scriptsize
\setlength{\tabcolsep}{1.5pt}		
\begin{center}
	\caption{The components of our methods based on DeFRCN. The AP$_{50}$ of DeFRCN on 1 to 10-shot of PASCAL VOC split1 and MS-COCO is reported.}
	\label{table:3}
	\begin{tabular}{lcccccccc}
		\hline\noalign{\smallskip}
		DeFRCN (Our impl.)~\cite{qiao2021defrcn}&Add&Crop&Select&CLIP& 1-shot & 3-shot &5-shot& 10-shot \\
		\noalign{\smallskip}
		\hline
		\noalign{\smallskip}
		\multirow{6}*{on PASCAL VOC} &&&&&52.5&60.8&62.7&61.9\\
		&\checkmark&&&&51.9&60&61.9&63.6\\
		&\checkmark&\checkmark&&&62.2&65.5&66.5&67.8\\
		&&&&\checkmark&61.8&66.6&66.3&67.4\\
		&\checkmark&\checkmark&\checkmark&&63.5&66.3&67.5&69\\
		&\checkmark&\checkmark&\checkmark&\checkmark&{\bf 67.5}&{\bf 69.8}&{\bf71.1}&{\bf71.5}\\
		\hline
		\noalign{\smallskip}
		\multirow{3}*{on MS-COCO}&&&&&16.2&25.4&28.9&33.6\\
		&\checkmark&\checkmark&&&25.4&28&30.1&33.8\\
		&\checkmark&\checkmark&\checkmark&&27&30&32.3&34.7\\
		&\checkmark&\checkmark&\checkmark&\checkmark&{\bf29.2}&{\bf32.2}&{\bf33.1}&{\bf35.1}\\
		\hline
	\end{tabular}
\end{center}
\vskip -0.1in
\end{table}

\subsection{Implementation Details}
Our baselines include DeFRCN~\cite{qiao2021defrcn} and MFDC~\cite{wu2022multi}, which are representative methods in FSOD. These methods use Faster R-CNN~\cite{ren2016faster} with  ResNet-101~\cite{he2016deep}, which are the same as almost all FSOD methods. The training strategies of our methods follow the selected baselines. The difference is that we add the generated novel images to the few-shot set for training. The $G$ in the setting of $(K+G)$-shot is 20. We adopt the ViT-B/32~\cite{radford2021learning} as the encoder of CLIP.

\subsection{Comparison with State-of-the-art Methods}

We compare our approaches to several competitive FSOD methods. The results are shown in Table~\ref{table:1} and Table~\ref{table:2}.  Following~\cite{han2021meta,li2021beyond,wu2020multi,cao2021few,kaul2022label}, we use a single run with the same training images to get the results of different shots.

\subsubsection{Results on PASCAL VOC and MS COCO.} 
Following~\cite{yan2019meta,wang2020frustratingly,sun2021fsce}, we provide the AP$_{50}$ of the novel classes on PASCAL VOC with three splits in Table~\ref{table:1}. By using synthetic data, our methods can outperform the baselines in all few-shot settings and achieve state-of-the-art performance. The most obvious improvement is up to 21.9$\%$. 

We report the COCO-style AP and AP$_{75}$ of the 20 novel classes on MS COCO in Table~\ref{table:2}. By using synthetic data, our methods can achieve state-of-the-art performance in almost all settings. The performance gain in the setting of 30-shot is low, since higher shots decline the influence of the synthetic images from another source which is different from the data from the current domain.

\subsection{Ablation Study}

Following~\cite{yan2019meta,sun2021fsce}, we do ablation studies on PASCAL VOC split 1. We use DeFRCN as the baseline.

\subsubsection{The Components in Our Pipeline}
There are four essential steps for applying synthetic data to FSOD: Add, Crop, Select, and CLIP. In Table~\ref{table:3}, Add copies the original synthetic images and pasts them onto the background. Crop means using saliency detection for the main target before the pasting operation. Select is finding the top 20 representative samples from the generated dataset. CLIP uses the encoders of CLIP after the few-shot model to filter the FP. We discover that cropping is crucial, which can achieve 4$\%$ to 10$\%$ AP$_{50}$ improvement in PASCAL VOC, since directly using the original synthetic data hurt the performance due to the inaccurate bounding boxes. Our selection strategy can improve the precision on most settings (up to 2$\%$ in PASCAL VOC). CLIP helps the model to filter most of FP and improve the performance up to 4$\%$. The best performance on MS COCO is also achieved by using all components.

\begin{figure}[t]
\vskip -0.1in
\begin{center}
	\centerline{\includegraphics[width=\columnwidth]{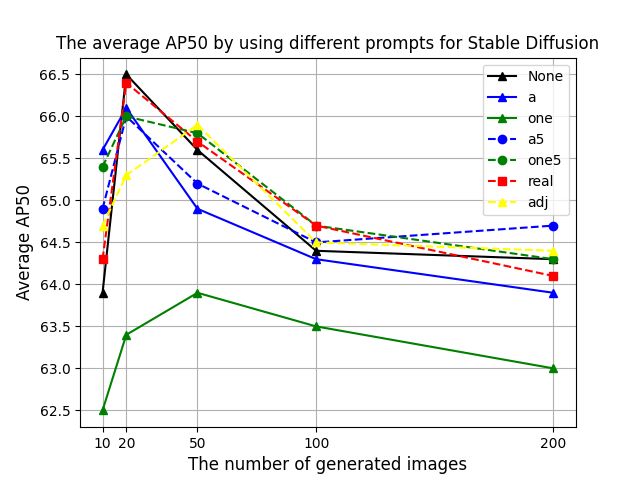}}
	\caption{The influence of different input text for Stable Diffusion and the number of synthetic data on the average AP$_{50}$ of PASCAL VOC split1 1 to 10-shot. There are seven types of text descriptions: None, a, a5, one, one5, adj, and real. The X-axis is the number of generated images ($G$) for each novel category.}
	\label{fig:prompt}
\end{center}
\vskip -0.2in
\end{figure}

\begin{table}[t]
\vskip -0.1in
\scriptsize
\setlength{\tabcolsep}{6pt}		
\begin{center}
	\caption{The type of several pasting methods. The results are averaged by using 10 to 200 generated images for training on 1 to 10-shot of PASCAL VOC split1. The saliency map is the result of saliency object detection. The segmentation map is the result of a supervised image segmentation method. Box uses the minimum enclosing rectangle of the saliency map.}
	\label{table:pasting}
	\begin{tabular}{lcccc}
		\hline\noalign{\smallskip}
		Methods& 1-shot & 3-shot &5-shot& 10-shot \\
		\noalign{\smallskip}
		\hline
		\noalign{\smallskip}
		Box &{\bf62.7}&{\bf65.2}&{\bf65.4}&{\bf66.8}\\
		Saliency map &60.4&63.1&63.9&65.5\\
		Segmentation map&61.5&63.9&64.2&66\\
		\hline
	\end{tabular}
\end{center}
\vskip -0.1in
\end{table}

\subsubsection{The Factors in Data Generation and Selection}
\label{sec:G}
There are three factors: the format of input text description, the number of used synthetic images (G in the setting of $(K+G)$-shot), and the type of pasting generated novel instances.

In Figure~\ref{fig:prompt}, we analyze the influence of the number of used synthetic images and the input text description for Stable Diffusion. As a result, we can get the following conclusions.
First, almost all input text formats perform best with 20 synthetic images.
Second, except for the format of "one", other formats have relatively high performance on average. In detail, "a5" and "one5" achieve the highest performance on average. "a5" has the most stable curve. Therefore, in other experiments, we use "a5" format with 20 synthetic images per class.
In Table~\ref{table:pasting}, we compare three different pasting methods: box, saliency map, and segmentation map, in which these types of novel instances are pasted onto the background. The saliency map uses the results of BASNet~\cite{qin2019basnet}. The segmentation map is from a supervised image segmentation method: deeplabv3+~\cite{chen2017rethinking}. Box uses the minimum enclosing rectangle of the saliency map. The results are averaged by using 10 to 200 generated images for training. In conclusion, the box type achieves the best performance in all settings.

\subsubsection{The Selection Module}

Table~\ref{table:sel} compares different synthetic data selection methods. All selection methods choose 20 samples out of 200 generated images per class. The features of samples (generated or original few-shot samples) are based on the image encoder of CLIP. 
We find that only considering the quality of synthetic data performs worse, such as CLIP$\_$max, since this operation reduces the diversity of data. Focusing on the high diversity of synthetic data can improve the few-shot model, such as using a clustering or uniform sampling method based on the CLIP score. The performance of Instance$\_$uniform is not satisfactory because there is a large deviation between the mean feature of few-shot samples and the actual prototype of a category. The best performance is achieved by spectral clustering.

\begin{table}[t]
\vskip -0.1in
\scriptsize
\setlength{\tabcolsep}{6pt}		
\begin{center}
	\caption{The comparison of different data selection methods. All selection methods choose 20 samples out of 200 generated images per class. The features of samples are obtained from the image encoder of CLIP. The results of the random selection are averaged by five different selections. Syn$\_$max selects the top 20 samples which have the minimum distance to the mean features of all generated images. CLIP$\_$max selects the samples with the top-20 CLIP score.
		Instance$\_$max selects the nearest neighbor samples according to the mean feature of few-shot instances.
		CLIP$\_$uniform is uniform sampling based on the CLIP score. Instance$\_$uniform means uniform sampling based on the distance between the mean features of the few-shot instances and the features of generated images. Kmeans$\_$cluster and Spectral$\_$cluster get 20 clusters for each category to select the nearest neighbor sample for each cluster. {\bf Black} is the best, and {\color{red}Red} is the second best.}
	\label{table:sel}
	\begin{tabular}{llcccc}
		\hline\noalign{\smallskip}
		Types&Methods& 1-shot & 3-shot &5-shot& 10-shot \\
		\noalign{\smallskip}
		\hline
		\noalign{\smallskip}
		\multirow{2}*{Others}&Random&62.6&65.9&66.8&68.2\\
		&Syn$\_$max&61.6&64.6&66.6&68.5\\
		\multirow{2}*{Quality-based}&CLIP$\_$max&59.9&64.2&65.9&67.2\\
		&Instance$\_$max&61.6&65.2&67.2&{\color{red}68.6}\\
		\noalign{\smallskip}
		\hline
		\noalign{\smallskip}
		\multirow{2}*{Sample-based}&CLIP$\_$uniform&{\color{red}63.4}&66.1&66.7&{\color{red}68.6}\\
		&Instance$\_$uniform&60.6&64.7&{\color{red}67.4}&68.3\\
		
		\multirow{2}*{Cluster-based}&Kmeans$\_$cluster&62.6&{\bf66.6}&66.6&67.9\\
		&Spectral$\_$cluster&{\bf63.5}&{\color{red}66.3}&{\bf67.5}&{\bf69}\\
		\hline
	\end{tabular}
\end{center}
\vskip -0.2in
\end{table}

\begin{figure}[t]
\vskip -0.1in
\begin{center}
	\centerline{\includegraphics[width=\columnwidth]{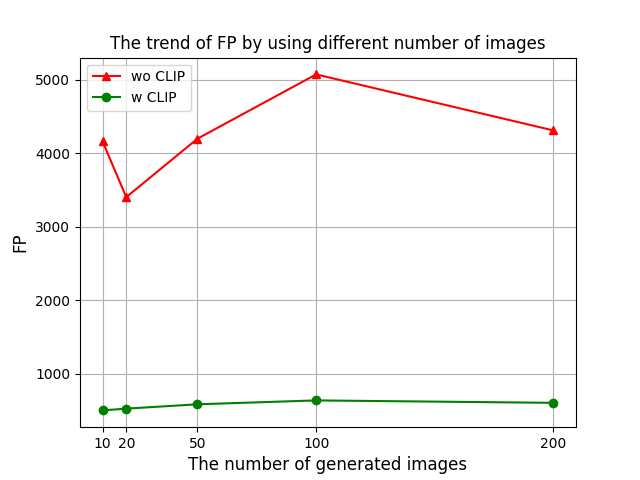}}
	\caption{The power of CLIP on reducing the number of FP. The generated images are added to the 1-shot setting of PASCAL VOC split1. w/o CLIP means the original few-shot model. w CLIP uses CLIP to filter the FP. The increase of the generated images used does not reduce the FP ratio. The FP ratio of using CLIP is much lower than that without CLIP.}
	\label{fig:CLIP-FP}
\end{center}
\vskip -0.3in
\end{figure}

\begin{table}[t]
\vskip -0.1in
\scriptsize
\setlength{\tabcolsep}{3pt}		
\begin{center}
	\caption{The filtering thresh for reducing the FP by using CLIP. The CLIP thresh is applied to CLIP score, which is lower than the CLIP thresh is defined as FP.}
	\label{table:CLIP}
	\begin{tabular}{lccccc}
		\hline\noalign{\smallskip}
		DeFRCN (Our impl.)~\cite{qiao2021defrcn}& Thresh&1-shot & 3-shot &5-shot& 10-shot\\
		\noalign{\smallskip}
		\hline
		\noalign{\smallskip}
		\multirow{5}*{CLIP Thresh}		 
		&0.5&65.2&68.4&68.1&69.6\\
		&0.3&65.8&68.6&67.5&69.2\\
		&0.2&66.3&{\bf70.2}&70&71.2\\
		&0.1&{\bf 67.5}&69.8&{\bf71.1}&{\bf71.5}\\
		&0&63.5&66.3&67.5&69\\
		\hline
	\end{tabular}
\end{center}
\vskip -0.1in
\end{table}

\begin{table}[t]
\scriptsize
\setlength{\tabcolsep}{3pt}		
\begin{center}
	\caption{The category list for  CLIP. The AP$_{50}$ of DeFRCN on 1 to 10-shot of PASCAL VOC split1 and MS-COCO is reported. The input of the text encoder of CLIP can be both the base and novel classes or the novel classes only. The experiments are conducted on PASCAL VOC and MS-COCO.}
	\label{table:CLIP-use}
	\begin{tabular}{lccccc}
		\hline\noalign{\smallskip}
		DeFRCN (Our impl.)~\cite{qiao2021defrcn}&Type& 1-shot & 3-shot &5-shot& 10-shot \\
		\noalign{\smallskip}
		\hline
		\noalign{\smallskip}
		\multirow{2}*{on PASCAL VOC} &On Novel&62.2&66.3&65.7&68.1\\
		&On Base+Novel&{\bf66.3}&{\bf70.2}&{\bf70}&{\bf71.2}\\
		\hline
		\noalign{\smallskip}
		\multirow{2}*{on MS-COCO}&On Novel&{\bf29.2}&{\bf32.2}&{\bf33.1}&{\bf35.1}\\
		&On Base+Novel&27.4&29.1&30.8&31.5\\
		\hline
	\end{tabular}
\end{center}
\vskip -0.2in
\end{table}

\subsubsection{How to Use CLIP}
\label{sec:CLIP}
In Figure~\ref{fig:CLIP-FP}, increasing the generated images can not reduce the FP ratio. The reduction of FP ratio by using CLIP is up to 90 $\%$. In Table~\ref{table:CLIP-use}, we compare two category lists used as the input of the text-encoder of CLIP: novel and base categories or only the novel categories. Using novel and base categories means that CLIP calculates the similarity between the input image and the input text of all categories(20 in PASCAL VOC and 80 in MS COCO). When using novel categories, CLIP calculates the similarity between the input image and the input text of novel categories(5 in PASCAL VOC and 20 in MS COCO). The results show that the optimal length of the name list is 20, which is all categories of PASCAL VOC and the novel categories of MS COCO.
Table~\ref{table:CLIP} shows the effect of CLIP thresh on performance. The detected objects with the CLIP score of the predicted categories, which are lower than the CLIP threshold, are defined as false positives. These predictions are removed for final evaluation. The optimal CLIP  threshold is 0.1.

\begin{figure}[t]
\vskip -0.1in
\begin{center}
	\centerline{\includegraphics[width=\columnwidth]{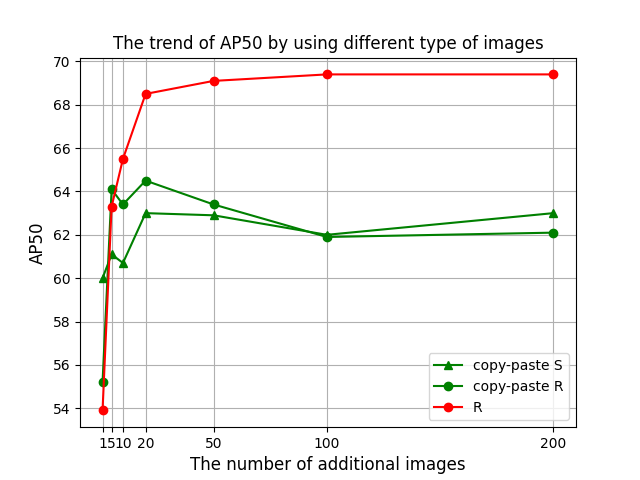}}
	\caption{The influence of different types of additional images per category: the images with the pasted synthetic novel instance (copy-paste S), the images with the pasted real novel instance (copy-paste R), and the real novel images (R). The results are obtained by adding these images to the original 1-shot set of PASCAL VOC split1 for training the DeFRCN.}
	\label{fig:upbound}
\end{center}
\vskip -0.3in
\end{figure}

\subsubsection{Other Analysis}

In Figure~\ref{fig:upbound}, we study the upper bound of the performance of using synthetic data. With the increase of synthetic data, copy-paste-based methods, such as pasting real novel instances or synthetic instances, have a lower upper bound than using the original novel images. We can notice that the gap between pasting real and synthetic instances is marginal, which means the synthetic images are ready for the FSOD tasks. Using more synthetic data can boost accuracy when the number of data is lower than 50, but the gain does not continue with the increase of generated data. The performance of FSOD is mainly limited by the copy-paste itself, not the synthetic data.

We compare another text-to-image generator (GLIDE) to Stable Diffusion. The results in Table~\ref{table:gen} show that the quality of the text-to-image generator affects the performance of FSOD.



\begin{table}[t]
\scriptsize
\setlength{\tabcolsep}{6pt}		
\begin{center}
	\caption{The comparison between different text-to-image generators: Stable Diffusion and GLIDE. The results of both methods are averaged by training five times. Twenty samples are randomly selected for each training.}
	\label{table:gen}
	\begin{tabular}{lcccc}
		\hline\noalign{\smallskip}
		Methods& 1-shot & 3-shot &5-shot& 10-shot \\
		\noalign{\smallskip}
		\hline
		\noalign{\smallskip}
		Stable Diffusion~\cite{rombach2022high} &{\bf62.6}&{\bf 65.9}&{\bf66.8}&{\bf68.2}\\
		GLIDE~\cite{nichol2021glide} &60.3&64.7&64.1&66.1\\
		\hline
	\end{tabular}
\end{center}
\vskip -0.2in
\end{table}

\section{Conclusion}

In this work, we leverage the external synthetic novel data from a text-to-image generator for FSOD. We focus on utilizing the synthetic novel data, including the copy-paste pipeline, the design of input text for Stable Diffusion, the number of used synthetic images, and the data selection strategy for discovering the representative synthetic data. Furthermore, we propose integrating CLIP into the FSOD model to solve the problem of high FP ratio in FSOD. Extensive experiments on the few-shot object detection datasets, i.e., PASCAL VOC and MS COCO, validate the effectiveness of our method.  

\section*{Acknowledgement} This work is sponsored Hetao Shenzhen-Hong Kong Science and Technology Innovation Cooperation Zone (HZQB-KCZYZ-2021045).

{\small
\bibliographystyle{ieee_fullname}
\bibliography{egbib}

\begin{thebibliography}{10}\itemsep=-1pt

\bibitem{cao2021few}
Yuhang Cao, Jiaqi Wang, Ying Jin, Tong Wu, Kai Chen, Ziwei Liu, and Dahua Lin.
\newblock Few-shot object detection via association and discrimination.
\newblock {\em Advances in Neural Information Processing Systems}, 34, 2021.

\bibitem{carion2020end}
Nicolas Carion, Francisco Massa, Gabriel Synnaeve, Nicolas Usunier, Alexander
  Kirillov, and Sergey Zagoruyko.
\newblock End-to-end object detection with transformers.
\newblock In {\em European conference on computer vision}, pages 213--229.
  Springer, 2020.

\bibitem{chen2017rethinking}
Liang-Chieh Chen, George Papandreou, Florian Schroff, and Hartwig Adam.
\newblock Rethinking atrous convolution for semantic image segmentation.
\newblock {\em arXiv preprint arXiv:1706.05587}, 2017.

\bibitem{dai2021dynamic}
Xiyang Dai, Yinpeng Chen, Bin Xiao, Dongdong Chen, Mengchen Liu, Lu Yuan, and
  Lei Zhang.
\newblock Dynamic head: Unifying object detection heads with attentions.
\newblock In {\em Proceedings of the IEEE/CVF conference on computer vision and
  pattern recognition}, pages 7373--7382, 2021.

\bibitem{dhariwal2021diffusion}
Prafulla Dhariwal and Alexander Nichol.
\newblock Diffusion models beat gans on image synthesis.
\newblock {\em Advances in Neural Information Processing Systems},
  34:8780--8794, 2021.

\bibitem{everingham2015pascal}
Mark Everingham, SM~Ali Eslami, Luc Van~Gool, Christopher~KI Williams, John
  Winn, and Andrew Zisserman.
\newblock The pascal visual object classes challenge: A retrospective.
\newblock {\em International journal of computer vision}, 111(1):98--136, 2015.

\bibitem{everingham2010pascal}
Mark Everingham, Luc Van~Gool, Christopher~KI Williams, John Winn, and Andrew
  Zisserman.
\newblock The pascal visual object classes (voc) challenge.
\newblock {\em International journal of computer vision}, 88(2):303--338, 2010.

\bibitem{fan2020few}
Qi Fan, Wei Zhuo, Chi-Keung Tang, and Yu-Wing Tai.
\newblock Few-shot object detection with attention-rpn and multi-relation
  detector.
\newblock In {\em Proceedings of the IEEE/CVF Conference on Computer Vision and
  Pattern Recognition}, pages 4013--4022, 2020.

\bibitem{ghiasi2021simple}
Golnaz Ghiasi, Yin Cui, Aravind Srinivas, Rui Qian, Tsung-Yi Lin, Ekin~D Cubuk,
  Quoc~V Le, and Barret Zoph.
\newblock Simple copy-paste is a strong data augmentation method for instance
  segmentation.
\newblock In {\em Proceedings of the IEEE/CVF Conference on Computer Vision and
  Pattern Recognition}, pages 2918--2928, 2021.

\bibitem{han2021query}
Guangxing Han, Yicheng He, Shiyuan Huang, Jiawei Ma, and Shih-Fu Chang.
\newblock Query adaptive few-shot object detection with heterogeneous graph
  convolutional networks.
\newblock In {\em Proceedings of the IEEE/CVF International Conference on
  Computer Vision}, pages 3263--3272, 2021.

\bibitem{han2021meta}
Guangxing Han, Shiyuan Huang, Jiawei Ma, Yicheng He, and Shih-Fu Chang.
\newblock Meta faster r-cnn: Towards accurate few-shot object detection with
  attentive feature alignment.
\newblock {\em arXiv preprint arXiv:2104.07719}, 2021.

\bibitem{han2022few}
Guangxing Han, Jiawei Ma, Shiyuan Huang, Long Chen, and Shih-Fu Chang.
\newblock Few-shot object detection with fully cross-transformer.
\newblock {\em arXiv preprint arXiv:2203.15021}, 2022.

\bibitem{hartigan1979algorithm}
John~A Hartigan and Manchek~A Wong.
\newblock Algorithm as 136: A k-means clustering algorithm.
\newblock {\em Journal of the royal statistical society. series c (applied
  statistics)}, 28(1):100--108, 1979.

\bibitem{he2016deep}
Kaiming He, Xiangyu Zhang, Shaoqing Ren, and Jian Sun.
\newblock Deep residual learning for image recognition.
\newblock In {\em Proceedings of the IEEE conference on computer vision and
  pattern recognition}, pages 770--778, 2016.

\bibitem{he2022synthetic}
Ruifei He, Shuyang Sun, Xin Yu, Chuhui Xue, Wenqing Zhang, Philip Torr, Song
  Bai, and Xiaojuan Qi.
\newblock Is synthetic data from generative models ready for image recognition?
\newblock {\em arXiv preprint arXiv:2210.07574}, 2022.

\bibitem{ho2020denoising}
Jonathan Ho, Ajay Jain, and Pieter Abbeel.
\newblock Denoising diffusion probabilistic models.
\newblock {\em Advances in Neural Information Processing Systems},
  33:6840--6851, 2020.

\bibitem{ho2022cascaded}
Jonathan Ho, Chitwan Saharia, William Chan, David~J Fleet, Mohammad Norouzi,
  and Tim Salimans.
\newblock Cascaded diffusion models for high fidelity image generation.
\newblock {\em J. Mach. Learn. Res.}, 23:47--1, 2022.

\bibitem{kang2019few}
Bingyi Kang, Zhuang Liu, Xin Wang, Fisher Yu, Jiashi Feng, and Trevor Darrell.
\newblock Few-shot object detection via feature reweighting.
\newblock In {\em Proceedings of the IEEE International Conference on Computer
  Vision}, pages 8420--8429, 2019.

\bibitem{kaul2022label}
Prannay Kaul, Weidi Xie, and Andrew Zisserman.
\newblock Label, verify, correct: A simple few shot object detection method.
\newblock In {\em Proceedings of the IEEE/CVF Conference on Computer Vision and
  Pattern Recognition}, pages 14237--14247, 2022.

\bibitem{li2021transformation}
Aoxue Li and Zhenguo Li.
\newblock Transformation invariant few-shot object detection.
\newblock In {\em Proceedings of the IEEE/CVF Conference on Computer Vision and
  Pattern Recognition}, pages 3094--3102, 2021.

\bibitem{li2021beyond}
Bohao Li, Boyu Yang, Chang Liu, Feng Liu, Rongrong Ji, and Qixiang Ye.
\newblock Beyond max-margin: Class margin equilibrium for few-shot object
  detection.
\newblock In {\em Proceedings of the IEEE/CVF Conference on Computer Vision and
  Pattern Recognition}, pages 7363--7372, 2021.

\bibitem{li2021few}
Yiting Li, Haiyue Zhu, Yu Cheng, Wenxin Wang, Chek~Sing Teo, Cheng Xiang,
  Prahlad Vadakkepat, and Tong~Heng Lee.
\newblock Few-shot object detection via classification refinement and
  distractor retreatment.
\newblock In {\em Proceedings of the IEEE/CVF Conference on Computer Vision and
  Pattern Recognition}, pages 15395--15403, 2021.

\bibitem{lin2023effective}
Shaobo Lin, Kun Wang, Xingyu Zeng, and Rui Zhao.
\newblock An effective crop-paste pipeline for few-shot object detection.
\newblock {\em arXiv preprint arXiv:2302.14452}, 2023.

\bibitem{lin2017feature}
Tsung-Yi Lin, Piotr Doll{\'a}r, Ross Girshick, Kaiming He, Bharath Hariharan,
  and Serge Belongie.
\newblock Feature pyramid networks for object detection.
\newblock In {\em Proceedings of the IEEE conference on computer vision and
  pattern recognition}, pages 2117--2125, 2017.

\bibitem{lin2014microsoft}
Tsung-Yi Lin, Michael Maire, Serge Belongie, James Hays, Pietro Perona, Deva
  Ramanan, Piotr Doll{\'a}r, and C~Lawrence Zitnick.
\newblock Microsoft coco: Common objects in context.
\newblock In {\em European conference on computer vision}, pages 740--755.
  Springer, 2014.

\bibitem{ng2001spectral}
Andrew Ng, Michael Jordan, and Yair Weiss.
\newblock On spectral clustering: Analysis and an algorithm.
\newblock {\em Advances in neural information processing systems}, 14, 2001.

\bibitem{nichol2021glide}
Alex Nichol, Prafulla Dhariwal, Aditya Ramesh, Pranav Shyam, Pamela Mishkin,
  Bob McGrew, Ilya Sutskever, and Mark Chen.
\newblock Glide: Towards photorealistic image generation and editing with
  text-guided diffusion models.
\newblock {\em arXiv preprint arXiv:2112.10741}, 2021.

\bibitem{park2022majority}
Seulki Park, Youngkyu Hong, Byeongho Heo, Sangdoo Yun, and Jin~Young Choi.
\newblock The majority can help the minority: Context-rich minority
  oversampling for long-tailed classification.
\newblock In {\em Proceedings of the IEEE/CVF Conference on Computer Vision and
  Pattern Recognition}, pages 6887--6896, 2022.

\bibitem{qiao2021defrcn}
Limeng Qiao, Yuxuan Zhao, Zhiyuan Li, Xi Qiu, Jianan Wu, and Chi Zhang.
\newblock Defrcn: Decoupled faster r-cnn for few-shot object detection.
\newblock In {\em Proceedings of the IEEE/CVF International Conference on
  Computer Vision}, pages 8681--8690, 2021.

\bibitem{qin2019basnet}
Xuebin Qin, Zichen Zhang, Chenyang Huang, Chao Gao, Masood Dehghan, and Martin
  Jagersand.
\newblock Basnet: Boundary-aware salient object detection.
\newblock In {\em Proceedings of the IEEE/CVF conference on computer vision and
  pattern recognition}, pages 7479--7489, 2019.

\bibitem{radford2021learning}
Alec Radford, Jong~Wook Kim, Chris Hallacy, Aditya Ramesh, Gabriel Goh,
  Sandhini Agarwal, Girish Sastry, Amanda Askell, Pamela Mishkin, Jack Clark,
  et~al.
\newblock Learning transferable visual models from natural language
  supervision.
\newblock In {\em International Conference on Machine Learning}, pages
  8748--8763. PMLR, 2021.

\bibitem{ramesh2022hierarchical}
Aditya Ramesh, Prafulla Dhariwal, Alex Nichol, Casey Chu, and Mark Chen.
\newblock Hierarchical text-conditional image generation with clip latents.
\newblock {\em arXiv preprint arXiv:2204.06125}, 2022.

\bibitem{redmon2017yolo9000}
Joseph Redmon and Ali Farhadi.
\newblock Yolo9000: better, faster, stronger.
\newblock In {\em Proceedings of the IEEE conference on computer vision and
  pattern recognition}, pages 7263--7271, 2017.

\bibitem{ren2016faster}
Shaoqing Ren, Kaiming He, Ross Girshick, and Jian Sun.
\newblock Faster r-cnn: Towards real-time object detection with region proposal
  networks.
\newblock {\em IEEE transactions on pattern analysis and machine intelligence},
  39(6):1137--1149, 2016.

\bibitem{rombach2022high}
Robin Rombach, Andreas Blattmann, Dominik Lorenz, Patrick Esser, and Bj{\"o}rn
  Ommer.
\newblock High-resolution image synthesis with latent diffusion models.
\newblock In {\em Proceedings of the IEEE/CVF Conference on Computer Vision and
  Pattern Recognition}, pages 10684--10695, 2022.

\bibitem{saharia2022photorealistic}
Chitwan Saharia, William Chan, Saurabh Saxena, Lala Li, Jay Whang, Emily
  Denton, Seyed Kamyar~Seyed Ghasemipour, Burcu~Karagol Ayan, S~Sara Mahdavi,
  Rapha~Gontijo Lopes, et~al.
\newblock Photorealistic text-to-image diffusion models with deep language
  understanding.
\newblock {\em arXiv preprint arXiv:2205.11487}, 2022.

\bibitem{song2020improved}
Yang Song and Stefano Ermon.
\newblock Improved techniques for training score-based generative models.
\newblock {\em Advances in neural information processing systems},
  33:12438--12448, 2020.

\bibitem{sun2021fsce}
Bo Sun, Banghuai Li, Shengcai Cai, Ye Yuan, and Chi Zhang.
\newblock Fsce: Few-shot object detection via contrastive proposal encoding.
\newblock In {\em Proceedings of the IEEE/CVF Conference on Computer Vision and
  Pattern Recognition}, pages 7352--7362, 2021.

\bibitem{wang2020frustratingly}
Xin Wang, Thomas~E Huang, Trevor Darrell, Joseph~E Gonzalez, and Fisher Yu.
\newblock Frustratingly simple few-shot object detection.
\newblock {\em arXiv preprint arXiv:2003.06957}, 2020.

\bibitem{wang2019meta}
Yu-Xiong Wang, Deva Ramanan, and Martial Hebert.
\newblock Meta-learning to detect rare objects.
\newblock In {\em Proceedings of the IEEE International Conference on Computer
  Vision}, pages 9925--9934, 2019.

\bibitem{wu2021universal}
Aming Wu, Yahong Han, Linchao Zhu, and Yi Yang.
\newblock Universal-prototype enhancing for few-shot object detection.
\newblock In {\em Proceedings of the IEEE/CVF International Conference on
  Computer Vision}, pages 9567--9576, 2021.

\bibitem{wu2020multi}
Jiaxi Wu, Songtao Liu, Di Huang, and Yunhong Wang.
\newblock Multi-scale positive sample refinement for few-shot object detection.
\newblock In {\em European Conference on Computer Vision}, pages 456--472.
  Springer, 2020.

\bibitem{wu2022multi}
Shuang Wu, Wenjie Pei, Dianwen Mei, Fanglin Chen, Jiandong Tian, and Guangming
  Lu.
\newblock Multi-faceted distillation of base-novel commonality for few-shot
  object detection.
\newblock In {\em Computer Vision--ECCV 2022: 17th European Conference, Tel
  Aviv, Israel, October 23--27, 2022, Proceedings, Part IX}, pages 578--594.
  Springer, 2022.

\bibitem{yan2019meta}
Xiaopeng Yan, Ziliang Chen, Anni Xu, Xiaoxi Wang, Xiaodan Liang, and Liang Lin.
\newblock Meta r-cnn: Towards general solver for instance-level low-shot
  learning.
\newblock In {\em Proceedings of the IEEE International Conference on Computer
  Vision}, pages 9577--9586, 2019.

\bibitem{zhang2022kernelized}
Shan Zhang, Lei Wang, Naila Murray, and Piotr Koniusz.
\newblock Kernelized few-shot object detection with efficient integral
  aggregation.
\newblock In {\em Proceedings of the IEEE/CVF Conference on Computer Vision and
  Pattern Recognition}, pages 19207--19216, 2022.

\bibitem{zhang2021hallucination}
Weilin Zhang and Yu-Xiong Wang.
\newblock Hallucination improves few-shot object detection.
\newblock In {\em Proceedings of the IEEE/CVF Conference on Computer Vision and
  Pattern Recognition}, pages 13008--13017, 2021.

\end{thebibliography}
}

\end{document}